\documentclass[10pt,twocolumn,letterpaper]{article}

\usepackage{cvpr}              % CAMERA-READY version
\usepackage{booktabs}

\usepackage[accsupp]{axessibility}  % Improves PDF readability for those with disabilities.

\definecolor{cvprblue}{rgb}{0.21,0.49,0.74}
\usepackage[pagebackref,breaklinks,colorlinks,allcolors=cvprblue]{hyperref}

\def\paperID{22}
\def\confName{CVPR}
\def\confYear{2026}

\title{Efficient INT8 Single-Image Super-Resolution via Deployment-Aware Quantization and Teacher-Guided Training}

\author{
Pham Phuong Nam Nguyen$^{1,3}$ \qquad
Nam Tien Le$^{2,3}$ \qquad
Thi Kim Trang Vo$^{1,3}$ \qquad
Nhu Tinh Anh Nguyen$^{2,3}$\\[0.6em]
$^1$University of Information Technology \\
$^2$Ho Chi Minh City University of Technology \\
$^3$Vietnam National University, Ho Chi Minh City
}

\begin{document}
\maketitle

\begin{abstract}
Efficient single-image super-resolution (SISR) requires balancing reconstruction fidelity, model compactness, and robustness under low-bit deployment, which is especially challenging for $\times 3$ SR. We present a deployment-oriented quantized SISR framework based on an extract--refine--upsample design. The student performs most computation in the low-resolution space and uses a lightweight re-parameterizable backbone with PixelShuffle reconstruction, yielding a compact inference graph. To improve quality without significantly increasing complexity, we adopt a three-stage training pipeline: Stage 1 learns a basic reconstruction mapping with spatial supervision; Stage 2 refines fidelity using Charbonnier loss, DCT-domain supervision, and confidence-weighted output-level distillation from a Mamba-based teacher; and Stage 3 applies quantization-aware training directly on the fused deploy graph. We further use weight clipping and BatchNorm recalibration to improve quantization stability. On the MAI 2026 Quantized 4K Image Super-Resolution Challenge test set, our final AIO\_MAI submission achieves 29.79 dB PSNR and 0.8634 SSIM, obtaining a final score of 1.8 under the target mobile INT8 deployment setting. Ablation on Stage 3 optimization shows that teacher-guided supervision improves the dynamic INT8 TFLite reconstruction from 29.91 dB / 0.853 to 30.0003 dB / 0.856, while the fixed-shape deployable INT8 TFLite artifact attains 30.006 dB / 0.857.
\end{abstract}

\section{Introduction}
\label{sec:intro}

Recent progress in deep learning and computer vision has led to substantial improvements across a wide range of tasks, including image retrieval \cite{nguyen2026itselfattentionguidedfinegrained,nguyen2025hybridunifiediterativenovel} and visual question answering \cite{nguyennhu2025stervlmspatiotemporalenhancedreference,nguyen2024improvinggeneralizationvisualreasoning,xiao2025medicalimages}. Among low-level vision tasks, single-image super-resolution (SISR), which seeks to reconstruct a high-resolution image from a low-resolution input, remains a fundamental and challenging problem. For a scale factor $\times 3$, the model must recover missing high-frequency details while preserving spatial fidelity. Although recent SR models achieve strong reconstruction fidelity, they typically do so by increasing representational capacity or architectural complexity. EDSR improves SR quality with a higher-capacity residual backbone \cite{lim2017edsr}; RCAN further deepens this design through residual-in-residual groups with channel attention for adaptive feature rescaling \cite{zhang2018rcan}; SwinIR replaces purely convolutional processing with residual Swin Transformer blocks to capture longer-range dependencies \cite{liang2021swinir}; and HAT further increases modeling complexity by combining hybrid attention and cross-window interaction for stronger texture reconstruction \cite{chen2023activating}. As a result, these models are harder to compress and less robust under low-bit deployment. This issue is especially critical in SR because small numerical errors can directly cause visible texture degradation and edge artifacts.
\begin{figure}[htbp]
\centering
\includegraphics[width=\linewidth]{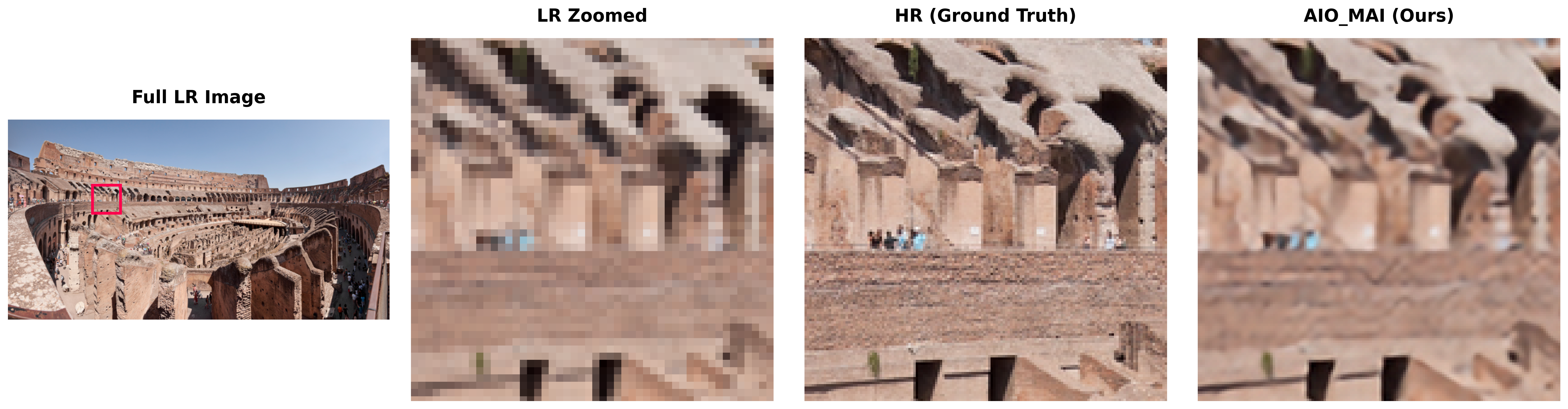} % <-- Nhớ đổi tên file ảnh ở đây
\caption{Overview of the proposed deployment-oriented quantized ×3 super-resolution framework. The method combines an LR-space extract–refine–upsample architecture, teacher-guided fidelity enhancement, and deploy-before-QAT optimization to achieve a strong trade-off between reconstruction quality, model compactness, and INT8 robustness. }
\label{fig:intro_teaser}
\end{figure}

A key step toward efficient SR is to keep most computation in the low-resolution (LR) space and postpone upsampling to the reconstruction stage. Methods such as ESPCN and LapSRN follow this principle and reduce unnecessary high-resolution computation while maintaining reconstruction quality \cite{shi2016espcn,lai2017lapsrn}. Later lightweight SR models, including CARN, IDN, IMDN, and RFDN, further improved the accuracy--efficiency trade-off through stronger feature reuse and selective information propagation \cite{ahn2018carn,hui2018idn,hui2019imdn,liu2020rfdn,phannguyen2025cycletrainingsemisuperviseddomain}. In parallel, structural re-parameterization provides another route to efficiency: RepVGG and MobileOne use expressive multi-branch training structures that can be fused into simple single-branch graphs for inference \cite{ding2021repvgg,vasu2023mobileone}. For SR, this is attractive because it improves optimization while keeping deployment simple and compact.

However, a lightweight design alone is not sufficient for robust low-bit SR. Quantization in SR is more difficult than in high-level vision because reconstruction quality is highly sensitive to activation range, rounding effects, and train--deploy mismatch. While QAT provides a strong basis for low-bit optimization \cite{jacob2018quantization}, recent SR-specific studies show that restoration features are highly dynamic and require tailored quantization strategies \cite{NEURIPS2023_QuantSR}. At the same time, compact SR models often lack the capacity to recover difficult textures and long-range structures. Knowledge distillation offers a natural way to transfer richer reconstruction priors from a stronger teacher to a lightweight student \cite{hinton2015distilling}. Recent Mamba-based restoration models, such as MambaIR and MambaIRv2, further show strong ability in modelling non-local dependencies for image restoration \cite{guo2024mambair,guo2025mambairv2}, making them suitable teachers for compact SR students.

% Motivated by these observations, we propose a compact and quantization-aware SISR framework for $\times 3$ super-resolution. The network follows an \emph{extract--refine--upsample} pipeline: a shallow input projection maps the LR image into feature space, a lightweight stack of MobileOne-style re-parameterizable blocks performs feature refinement entirely in LR space, and a PixelShuffle head reconstructs the HR output with a global skip pathway to preserve coarse structure. To optimize both reconstruction fidelity and low-bit robustness, training is conducted in three stages. Stage 1 initializes a stable LR-to-HR mapping with L1 supervision. Stage 2 refines the student using a robust Charbonnier loss \cite{lai2017lapsrn}, DCT-domain frequency supervision \cite{cai2021freqnet}, and confidence-weighted output-level distillation \cite{hinton2015distilling,zhang2022camkd} from a pretrained MambaIRv2Light teacher \cite{guo2025mambairv2}. Stage 3 applies QAT directly on the fused deploy graph \cite{jacob2018quantization}; clipping-aware quantization control and weight clipping are used to suppress quantization noise \cite{choi2018pact,sakr2022octav}, and BatchNorm recalibration is performed afterward to restore consistent activation statistics before quantized inference \cite{shomron2020bnrecal}.

Motivated by the above observations, we propose a compact and quantization-aware SISR framework for $\times 3$ super-resolution. Our network adopts an \emph{extract--refine--upsample} design: a shallow input stem projects the LR image into feature space, a sequence of lightweight MobileOne-style re-parameterizable blocks refines features entirely in the LR domain, and a PixelShuffle reconstruction head produces the HR output with a global skip pathway for coarse-structure preservation. 

Training is organized into three stages to align fidelity optimization with low-bit deployment. Stage~1 learns a stable base mapping using $L_{1}$ reconstruction. Stage~2 improves reconstruction quality with a hybrid objective consisting of robust Charbonnier loss \cite{lai2017lapsrn}, DCT-domain frequency supervision for explicit high-frequency constraint \cite{cai2021freqnet}, and confidence-weighted output-level distillation \cite{hinton2015distilling,zhang2022camkd} from a pretrained MambaIRv2Light teacher \cite{guo2025mambairv2}. Stage 3 applies QAT directly on the fused deploy graph \cite{jacob2018quantization}; clipping-based weight control and BatchNorm recalibration are introduced to improve quantization stability before low-bit inference \cite{shomron2020bnrecal}. Together, these components couple efficient LR-space reconstruction with teacher-guided fidelity enhancement and deployment-aware quantization.

Rather than introducing a new SR primitive, our contribution lies in a deployment-oriented INT8 super-resolution pipeline that tightly aligns training-time optimization with the actual deployed inference graph.

In summary, our method is based on the idea that efficient low-bit SR requires joint optimization of architecture, supervision, and deployment consistency. The main contributions are:
\begin{itemize}
    \item A compact LR-space $\times 3$ SR architecture with MobileOne-style re-parameterizable refinement and PixelShuffle reconstruction.
    
    \item A fidelity-oriented teacher--student training strategy combining spatial reconstruction, DCT-domain supervision, and confidence-weighted distillation from a Mamba-based teacher.
    
    \item A deployment-aware quantization pipeline that performs QAT on the fused deploy graph with weight clipping and BatchNorm recalibration for improved low-bit robustness.
    
    \item On the MAI 2026 Quantized 4K Image Super-Resolution Challenge test set, our final AIO\_MAI submission achieves 29.79 dB PSNR and 0.8634 SSIM, obtaining a final score of 1.8 under the target mobile INT8 deployment setting.
    
    \item Ablation on backbone design shows that the MobileOne-style block provides the strongest FP32/INT8 performance and the smallest quantization drop among the tested building blocks, outperforming both RepConv and our re-parameterizable depthwise block (RepDW).
    
    \item Ablation on Stage 3 optimization shows that adding teacher-guided supervision improves INT8 reconstruction from 29.91 dB / 0.853 to 30.0003 dB / 0.856, while the exported TFLite model preserves this gain at 30.006 dB / 0.857.
\end{itemize}

\section{Related Works}
\label{sec:formatting}

\paragraph{From high-capacity SR to efficient LR-space reconstruction.}
Early super-resolution (SR) models mainly improved fidelity by increasing network depth and capacity. EDSR is a representative high-capacity residual model \cite{lim2017edsr}. To reduce computation, later methods moved most feature extraction to the low-resolution (LR) space and postponed upsampling to the reconstruction head. ESPCN used sub-pixel convolution to reconstruct directly from LR features \cite{shi2016espcn}, while LapSRN further improved efficiency with progressive coarse-to-fine residual reconstruction \cite{lai2017lapsrn}. These works established the main principle of efficient SR: keep computation in LR space and upsample only at the end.

\paragraph{Lightweight SR via feature distillation.}
Once LR-space reconstruction became standard, efficient SR research focused on improving feature reuse under limited budgets. CARN enhanced information flow through cascading connections \cite{ahn2018carn}. IDN introduced information distillation blocks to jointly refine and compress features \cite{hui2018idn}. IMDN extended this design with progressive multi-distillation and selective fusion \cite{hui2019imdn}, and RFDN further simplified distilled feature propagation with lightweight connections and shallow residual blocks \cite{liu2020rfdn}. Overall, this line evolved toward more explicit feature selection and reuse in compact SR backbones.

\paragraph{Structural re-parameterization for efficient inference.}
Another route to efficiency is to separate train-time expressiveness from inference-time simplicity. RepVGG showed that multi-branch training blocks can be analytically fused into a plain convolutional network for deployment \cite{ding2021repvgg}. MobileOne extended this idea with a more lightweight over-parameterized design that is folded into single-branch operators at inference \cite{vasu2023mobileone}. For SR, such designs are attractive because they preserve stronger optimization during training while producing a compact and regular inference graph.

\paragraph{State-space modelling and Mamba-based restoration.}
Recent restoration models have explored state-space architectures to model longer-range dependencies more effectively than purely local convolutions. MambaIR adapts Mamba to low-level restoration with local enhancement and channel attention \cite{guo2024mambair}. MambaIRv2 further strengthens this line with attentive state-space modelling for better non-local interaction \cite{guo2025mambairv2}. For efficient SR, DVMSR shows that Vision Mamba can be combined with a lightweight design and distillation \cite{lei2024dvmsr}. This suggests that compact SR can benefit from both efficient local processing and stronger global context modelling.

\paragraph{Knowledge distillation for compact SR.}
Knowledge distillation transfers richer reconstruction behaviour from a strong teacher to a lightweight student. Hinton \emph{et al.} introduced distillation as a general soft-supervision framework for model compression \cite{hinton2015distilling}. In SR, this is especially useful because compact models often struggle with fine textures and long-range consistency. Recent works such as DVMSR confirm that distillation can improve lightweight SR quality without increasing inference complexity \cite{lei2024dvmsr}.

\paragraph{Quantization for efficient super-resolution.}
Quantization is essential for efficient SR, but SR is highly sensitive to quantization error because quality is measured at the pixel level. Jacob \emph{et al.} established the standard integer-only QAT framework \cite{jacob2018quantization}.  Later SR-specific methods showed that restoration features are highly dynamic and need tailored calibration. Tu \emph{et al.} addressed this with SR-oriented calibration and clipping \cite{tu2023ptqsr}, while QuantSR further improved low-bit SR with SR-aware quantization modules \cite{NEURIPS2023_QuantSR}. Recent challenge reports on quantized SR for mobile NPUs further show that practical low-bit SR must also satisfy full INT8 compatibility, hardware-friendly operators, and real-device efficiency constraints \cite{ignatov2025maiqsr,ignatov2022mobileai}. This reflects the shift from general low-bit methods to SR-aware and deployment-aware quantization.

\paragraph{Summary.}
Overall, the literature has evolved from efficient LR-space reconstruction to lightweight feature distillation, re-parameterized backbones, and SR-aware quantization. More recently, Mamba-based restoration, teacher--student SR, and mobile-oriented challenge reports have emphasized stronger global modeling and end-to-end deployability under strict INT8 constraints \cite{ignatov2025maiqsr,ignatov2022mobileai}. Our work follows this trend by combining efficient SR design, compact backbone construction, teacher-guided supervision, and deployment-aware quantization.

\section{Methodology }
\label{sec:method}

\subsection{Data Preprocessing}

We address single-image super-resolution (SISR) at scale factor $\times 3$, where a low-resolution RGB image $x \in \mathbb{R}^{H \times W \times 3}$ is mapped to a high-resolution prediction $\hat{y} \in \mathbb{R}^{3H \times 3W \times 3}$. Training uses paired DIV2K LR--HR images, while validation is performed on full-resolution pairs without border shaving.

% Checkpoints are selected by full-image RGB PSNR, and SSIM is reported as a secondary metric \cite{ignatov2025maiqsr}.

All images are stored in RGB space with intensity range $[0,255]$, while losses are computed in the normalized domain:
\begin{equation}
y_{01}=\frac{y}{255}, \qquad \hat{y}_{01}=\frac{\hat{y}}{255}.
\end{equation}

During training, aligned LR--HR patches are randomly cropped. For an LR crop of size $p \times p$ with top-left coordinate $(u,v)$, the corresponding HR crop is
\begin{equation}
y\bigl[3v:3(v+p),\;3u:3(u+p)\bigr].
\end{equation}
The LR patch size is $128 \times 128$ in Stage~1, $160 \times 160$ in Stage~2, and $144 \times 144$ in Stage~3. We also apply random horizontal flips, vertical flips, and transpose augmentation.

\begin{figure*}[!t]
    \centering
    \includegraphics[width=\textwidth,height=0.32\textheight,keepaspectratio]{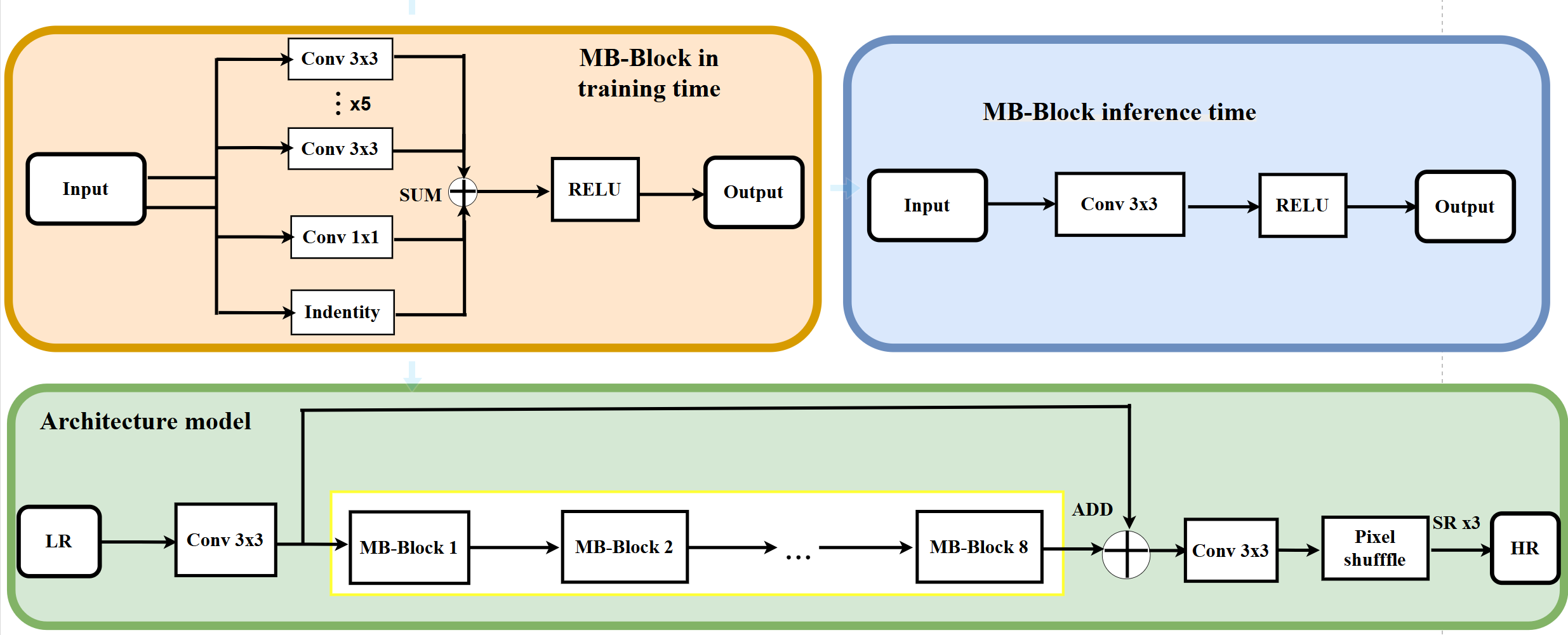}
    \caption{Architecture of the proposed student network. A shallow input stem first projects the LR image into feature space, followed by a stack of MobileOne-style re-parameterizable blocks operating entirely in the LR domain. A global feature skip preserves coarse structures, and a PixelShuffle head reconstructs the final ×3 HR output. At deployment, all training-time branches are fused into a compact single-branch inference graph.}
    \label{fig:placeholder}
\end{figure*}

\subsection{Network Architecture}

Our student network follows an extract--refine--upsample design for deployment-constrained SR. Most computation is performed in LR space and upsampling is deferred to the final PixelShuffle layer \cite{shi2016espcn}. To improve train-time capacity while keeping inference efficient, we use MobileOne-style structural re-parameterization \cite{ding2021repvgg,vasu2023mobileone}.

The LR input is first projected by a $3 \times 3$ convolution:
\begin{equation}
f_0=\phi_{\mathrm{in}}(x),
\end{equation}
where $f_0 \in \mathbb{R}^{H \times W \times C}$ and $C=32$.

The backbone contains $N=8$ MobileOne-style blocks. Each block uses four $3 \times 3$ convolution branches, one $1 \times 1$ branch, one identity branch, BatchNorm on active branches, and a final ReLU:
\begin{equation}
\begin{aligned}
B(f)=\sigma\!\Big(
&\sum_{i=1}^{5}\mathrm{BN}_i\!\left(\mathrm{Conv}^{(i)}_{3\times3}(f)\right) \\
&+ \mathrm{BN}_{1\times1}\!\left(\mathrm{Conv}_{1\times1}(f)\right)
+ \mathrm{BN}_{\mathrm{id}}(f)
\Big),
\end{aligned}
\end{equation}
where $\sigma(\cdot)=\mathrm{ReLU}(\cdot)$.

After training, all branches are fused into a single equivalent $3 \times 3$ convolution. For a convolution-BN branch with parameters $(W,b)$ and BN statistics $(\mu,\sigma^2,\gamma,\beta)$, the fused parameters are
\begin{equation}
\widetilde{W}=\frac{\gamma}{\sqrt{\sigma^2+\epsilon}}\,W,
\qquad
\widetilde{b}=\beta+\frac{\gamma}{\sqrt{\sigma^2+\epsilon}}(b-\mu).
\end{equation}

A feature-level global skip is applied after the backbone:
\begin{equation}
f=f_N+f_0.
\end{equation}
The refined feature map is projected to $27$ channels and upsampled by PixelShuffle:
\begin{equation}
z=\phi_{\mathrm{out}}(f), \qquad
\hat{y}=\mathrm{PixelShuffle}_3(z).
\end{equation}

The final model uses $8$ MobileOne-style blocks and $32$ channels, without an image-residual branch. Because the backbone operates in LR space, the main computation scales with $H \times W$ rather than $3H \times 3W$, which is favourable for mobile deployment \cite{shi2016espcn,ignatov2025maiqsr}.

\subsection{Training Objectives}

The final run uses a compact objective set: L1 reconstruction in Stage~1, and Charbonnier reconstruction with DCT supervision and output-level knowledge distillation (KD) in Stages~2 and~3. Specifically,

In Stage~1, we optimize:
\begin{equation}
\mathcal{L}_{\mathrm{L1}}=
\frac{1}{N}\left\lVert \hat{y}_{01}-y_{01}\right\rVert_1.
\end{equation}

In Stages~2 and~3, L1 is replaced by the Charbonnier loss \cite{lai2017lapsrn}:
\begin{equation}
\mathcal{L}_{\mathrm{char}}=
\frac{1}{N}\sum_i
\sqrt{\left(\hat{y}^{\,i}_{01}-y^{\,i}_{01}\right)^2+\epsilon^2},
\end{equation}
where $\epsilon=10^{-3}$.

To improve frequency fidelity, we also use
\begin{equation}
\mathcal{L}_{\mathrm{DCT}}=
\left\lVert D(\hat{y}_{01})-D(y_{01})\right\rVert_1.
\end{equation}

For output-level KD, a pretrained MambaIRv2Light $\times 3$ model provides teacher predictions $t$ \cite{guo2025mambairv2}:
\begin{equation}
\mathcal{L}_{\mathrm{KD}}=
\frac{1}{N}\sum_p w(p)\left|\hat{y}_{01}(p)-t(p)\right|.
\end{equation}
The confidence weight is computed from teacher error:
\begin{equation}
e(p)=\frac{1}{3}\sum_{c=1}^{3}\left|t_c(p)-y_{01,c}(p)\right|,
\end{equation}
\begin{equation}
w(p)=\mathrm{clip}\!\left(\exp(-\gamma e(p)),\,w_{\min},\,w_{\max}\right),
\end{equation}
with $\gamma=10.0$, $w_{\min}=0.10$, and $w_{\max}=0.75$.

The stage-wise objectives are
\begin{equation}
\mathcal{L}^{(1)}=\mathcal{L}_{\mathrm{L1}},
\end{equation}
\begin{equation}
\mathcal{L}^{(2)}=
\mathcal{L}_{\mathrm{char}}
+0.02\,\mathcal{L}_{\mathrm{DCT}}
+0.03\,\mathcal{L}_{\mathrm{KD}},
\end{equation}
and
\begin{equation}
\mathcal{L}^{(3)}=
\mathcal{L}_{\mathrm{char}}
+0.015\,\mathcal{L}_{\mathrm{DCT}}
+\lambda_{\mathrm{KD}}(t)\mathcal{L}_{\mathrm{KD}}.
\end{equation}

\begin{figure*}[!t]
    \centering
   \includegraphics[width=\textwidth,height=0.32\textheight,keepaspectratio]{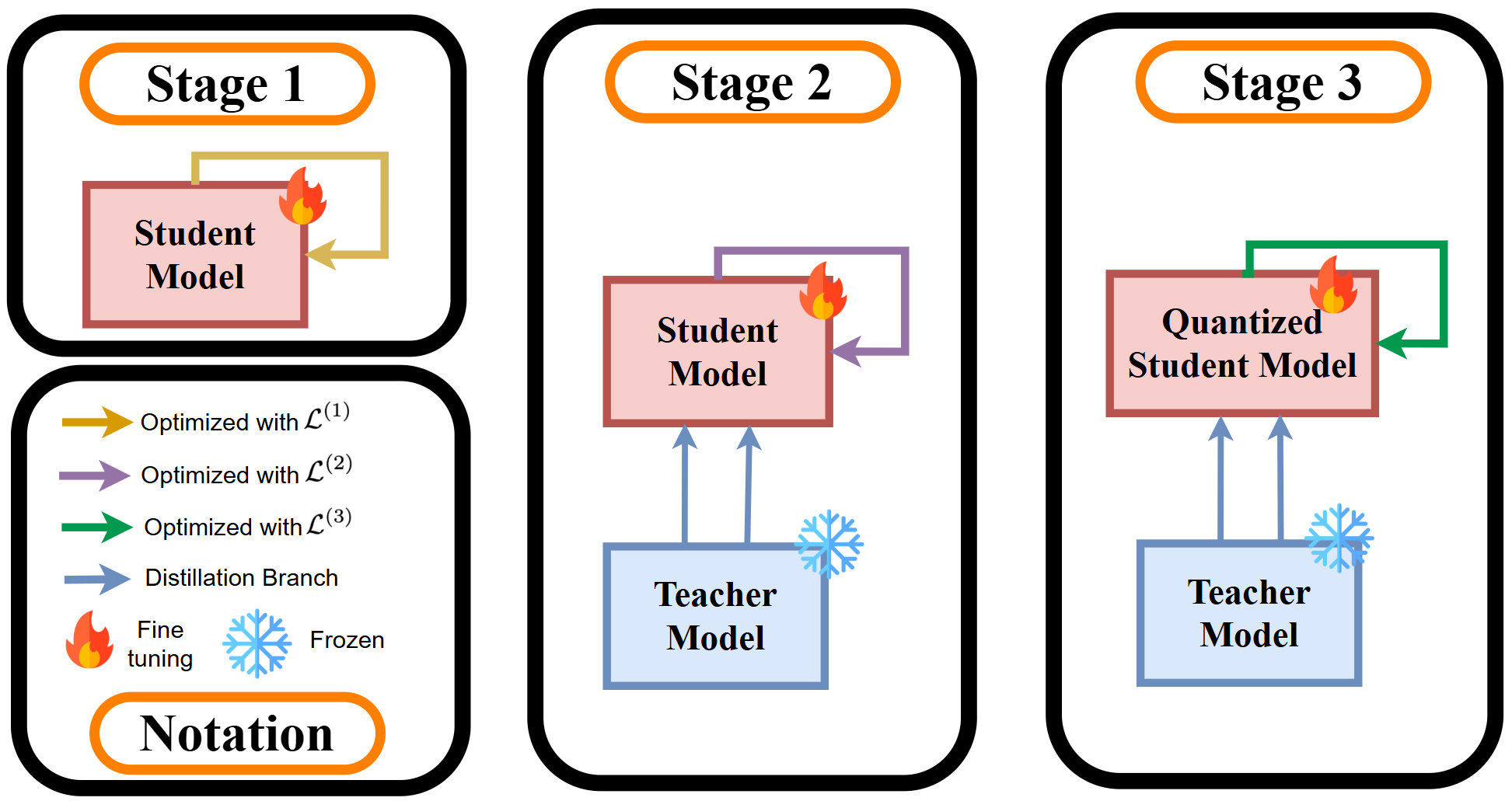}
    \caption{Overall training and deployment pipeline. Stage 1 learns a stable spatial mapping using L1 reconstruction, Stage 2 improves fidelity with Charbonnier loss, DCT-domain supervision, and teacher-guided output distillation, and Stage 3 performs QAT directly on the fused deploy graph before TFLite export. This staged design reduces the train–deploy mismatch and improves low-bit stability.}
    \label{fig:placeholder}
\end{figure*}

\subsection{Deployment-Aware Quantization}

\paragraph{The Quantization Mismatch Problem.} 
To deploy the proposed super-resolution model on mobile Neural Processing Units (NPUs) with highly constrained integer arithmetic, we quantize the network to 8-bit precision \cite{jacob2018quantization}. However, a significant challenge in quantizing structurally re-parameterized networks \cite{ding2021repvgg,vasu2023mobileone} is the discrepancy between the multi-branch training topology and the plain single-branch inference graph. Applying Quantization-Aware Training (QAT) directly to the multi-branch structure causes severe performance degradation when branches are analytically fused, as quantization errors compound unpredictably.

\paragraph{Deploy-before-QAT Strategy.} 
To alleviate this structural mismatch, we propose a Deploy-before-QAT paradigm. Prior to the initialization of quantization-aware training, we recalibrate the BatchNorm statistics using forward-only training mini-batches to ensure accurate running estimates. Subsequently, the multi-branch blocks are fully collapsed into single-branch $3\times3$ convolutions. The QAT operators, specifically utilizing PyTorch FX graph-mode tracing with the QNNPACK backend, are inserted directly into this fused graph. This ensures a strict mathematical alignment between the simulated fake-quantization during training and the actual integer execution during deployment.

\paragraph{Three-Phase QAT Curriculum.} 
The quantized network is optimized following a rigorous three-phase curriculum designed to progressively transition the model into a fully quantized state. During the initial phase, spanning the first 30 epochs, both the network weights and the quantization observers are active, allowing the model to adapt to the quantization constraints while continuously calibrating the scale and zero-point parameters. In the second phase, from epoch 30 to 90, the observers are disabled to freeze the quantization grid, compelling the network to fine-tune its weights under strict and unchanging 8-bit integer boundaries via the straight-through estimator (STE). Finally, from epoch 90 to 150, the fake-quantization nodes are completely frozen, and the network purely minimizes residual errors under the exact conditions of the target integer arithmetic. Upon completion, the optimized QAT graph is seamlessly exported to a TFLite flatbuffer, where graph operations are mapped to native 8-bit integer operators and tensors are converted to the NHWC layout, ensuring optimal hardware compatibility and a small precision drop in practice.
% Fallback in case placeins is not loaded in the main preamble
\providecommand{\FloatBarrier}{}

\section{Experimental Results}

\subsection{Dataset}
We follow the dataset setting of the Mobile AI 4K Quantized Image Super-Resolution Challenge and use DIV2K for the $\times 3$ single-image super-resolution task \cite{ignatov2025maiqsr}. DIV2K is a widely used high-resolution restoration benchmark consisting of 1000 RGB images with 2K resolution, including 800 images for training, 100 for validation, and 100 for testing \cite{agustsson2017div2k}. The dataset covers diverse content, including human subjects, man-made objects, urban scenes, plants, animals, and natural environments, with high visual quality and limited degradations.

\subsection{Experimental Setup}
Training follows a three-stage curriculum that progressively aligns floating-point reconstruction with low-bit deployment constraints \cite{jacob2018quantization,ignatov2025maiqsr}. The student first learns a stable spatial mapping, then improves reconstruction fidelity under stronger supervision, and finally adapts to the numerical constraints of INT8 inference.

\textbf{Stage 1.}
The floating-point student is trained for 600 epochs with initial learning rate $10^{-3}$ and LR patch size $128 \times 128$ using Adam, cosine warmup, and the L1 objective, providing a stable initialization before stronger fidelity-oriented supervision is introduced.

\textbf{Stage 2.}
We fine-tune for 200 epochs with initial learning rate $3\times10^{-5}$ and LR patch size $160 \times 160$. The loss combines Charbonnier, DCT, and output-level KD, with gradient clipping (max norm $1.0$), EMA (decay $0.999$), and post-update weight clipping. This stage focuses on texture and frequency recovery under a larger local context. Before Stage~3, the best checkpoint undergoes BatchNorm recalibration with 64 forward-only mini-batches and is fused into deploy form so that quantized training matches the inference graph.

\textbf{Stage 3.}
We perform QAT for 150 epochs with initial learning rate $10^{-6}$ and LR patch size $144 \times 144$. QAT is applied directly to the fused deploy graph rather than the multi-branch training graph, reducing the mismatch between fake-quantized training and deployment inference \cite{jacob2018quantization}. We use FX graph-mode QAT in PyTorch with the QNNPACK backend; observers are disabled after epoch 30, fake quantization is frozen after epoch 90, mixed precision is disabled, and the effective batch size is 1. This schedule allows the model to first adapt to quantization and then converge under fixed quantization boundaries that better reflect actual INT8 deployment.

During Stages~2 and~3, weights are clipped after each optimization step:
\begin{equation}
W \leftarrow \mathrm{clip}(W,W_{\min},W_{\max}).
\end{equation}
This operation reduces extreme weight excursions and improves stability in the low-bit setting. For deployment on mobile NPUs, the fused PyTorch QAT model is exported to TFLite with minimal graph transformation, including NCHW-to-NHWC conversion and mapping of fake-quantization nodes to native INT8 operators.

\textbf{Validation.}
Validation is performed on full RGB images without border shaving. Full-image RGB PSNR is used for checkpoint selection, while SSIM with an $11 \times 11$ Gaussian window and $\sigma=1.5$ is reported as a secondary metric. We adopt full-image RGB evaluation because it better reflects practical deployment behavior than cropped or luminance-only reporting.

Unless otherwise specified, Stage-3 ablations are reported on the dynamic fully-quantized INT8 TFLite artifact, which we use as the primary fidelity proxy. By contrast, deploy/runtime comparisons are reported on the fixed-shape deployable INT8 TFLite artifact. We additionally use a dynamic floating-point TFLite model only as a reference for consistency analysis.

\FloatBarrier

\subsection{Quantitative Results}

As shown in Table~\ref{tab:mai_results}, the proposed method achieves competitive performance under the target INT8 mobile deployment setting, indicating a favorable trade-off between reconstruction fidelity and runtime efficiency under practical hardware constraints.

\begin{table*}[t]
\centering
\caption{\textbf{Overall results on the MAI 2026 Quantized 4K Image Super-Resolution Challenge. Our method achieves a competitive balance between reconstruction fidelity and runtime efficiency under the target INT8 mobile deployment setting.}}
\label{tab:mai_results}
\renewcommand{\arraystretch}{1.08}
\resizebox{0.84\textwidth}{!}{%
\begin{tabular}{lcccccccc}
\toprule
\textbf{Team} & \multicolumn{2}{c}{\textbf{FP32}} & \multicolumn{2}{c}{\textbf{INT8}} & \multicolumn{3}{c}{\textbf{Runtime (ms)}} & \textbf{Final Score} \\
\cmidrule(lr){2-3} \cmidrule(lr){4-5} \cmidrule(lr){6-8}
~ & \textbf{PSNR} & \textbf{SSIM} & \textbf{PSNR} & \textbf{SSIM} & \textbf{CPU} & \textbf{GPU} & \textbf{NPU} & ~ \\
\midrule
AntYSP       & 30.04 & 0.8757 & 29.96 & 0.8729 & 338    & 50.1  & 4.33          & 21.8 \\
AntSR        & 30.08 & 0.8764 & 29.98 & 0.8731 & 404    & 58.5  & 4.52          & 21.5 \\
z6           & N.A.  & N.A.   & 29.93 & 0.8699 & 330    & 50.7  & 5.49          & 16.5 \\
koreatechV2  & N.A.  & N.A.   & 29.06 & 0.8499 & 464    & 69.8  & 4.35          & 6.2 \\
IN2GM        & 29.88 & 0.8708 & 29.85 & 0.8701 & 526    & 58.6  & 46.5          & 1.7 \\
LoongSR      & 30.43 & 0.8831 & 29.76 & 0.8586 & 5729   & 5557  & \textit{N/C}  & 0.01 \\
Godzilla     & N.A.  & N.A.   & 30.39 & 0.8831 & 215160 & OOM   & \textit{N/C}  & 0.0008 \\
Aimf\_SR     & 27.68 & 0.7923 & 10.46 & 0.2580 & 10806  & Error & \textit{N/C}  & 0 \\
IVCL         & 30.34 & 0.8814 & N.A.  & N.A.   & \textit{Parsing} & Error & \textit{N/C} & error \\
\midrule
\textbf{AIO\_MAI (OURS)} & 29.98 & 0.8730 & 29.79 & 0.8634 & 610 & 114 & 41.1 & 1.8 \\
\bottomrule
\end{tabular}%
}
\renewcommand{\arraystretch}{1.0}
\end{table*}

A key observation is that this performance does not come from scaling model size or increasing architectural complexity. Instead, it comes from combining LR-space computation, a lightweight re-parameterizable backbone, and fused-graph QAT, which together make the method more suitable for mobile NPU deployment. This suggests that careful coordination between architecture and deployment-aware optimization can be more effective than relying on larger models alone.

We compare our method with Bicubic, FSRCNN, and ABPN to assess whether a compact deployment-oriented model can remain competitive with established SR baselines. As shown in Table~\ref{tab:comparison}, the fixed-shape deployable INT8 TFLite artifact achieves 30.13 dB PSNR and 0.858 SSIM, clearly outperforming Bicubic and FSRCNN. Compared with ABPN, it attains nearly identical PSNR (30.13 vs. 30.15) while achieving higher SSIM (0.858 vs. 0.852), indicating better preservation of visually important structural patterns under quantized deployment. Before deployment, the Stage 2 FP32 model reaches 30.28 dB PSNR and 0.863 SSIM, the best results among all compared methods. The small drop from FP32 to deployable INT8 further demonstrates the robustness of our architecture and optimization pipeline for low-bit deployment. Overall, the results confirm that the proposed method maintains a favorable trade-off between efficiency and quality despite its very small parameter budget.

\begin{table}[htbp]
\centering
\caption{Comparison with standard super-resolution baselines. Results for our method are reported using the fixed-shape deployable INT8 TFLite artifact. Despite using a compact deployable model with only 82K parameters, the proposed method achieves stronger structural similarity than larger or less deployment-oriented baselines.}
\label{tab:comparison}
\renewcommand{\arraystretch}{1.08}
\resizebox{0.96\linewidth}{!}{%
\begin{tabular}{lccc}
\toprule
\textbf{Method} & \textbf{DataType} & \textbf{PSNR (dB)} & \textbf{SSIM} \\
\midrule
Bicubic & FP32 & 28.26 & 0.828 \\
FSRCNN \cite{dong2016acceleratingsuperresolutionconvolutionalneural} & FP32 & 29.45 & 0.838 \\
ABPN \cite{du2021anchorbased} & INT8 & \underline{30.15} & 0.852 \\
\midrule
\textbf{Ours (Stage 2)} & FP32 & \textbf{30.28} & \textbf{0.863} \\
\textbf{Ours (Deploy)} & INT8 & 30.13 & \underline{0.858} \\
\bottomrule
\end{tabular}%
}
\renewcommand{\arraystretch}{1.0}
\end{table}

\FloatBarrier

\subsection{Qualitative Results}
\label{subsec:qualitative}

\begin{figure*}[!t]
\centering
\includegraphics[width=\textwidth,height=0.32\textheight,keepaspectratio]{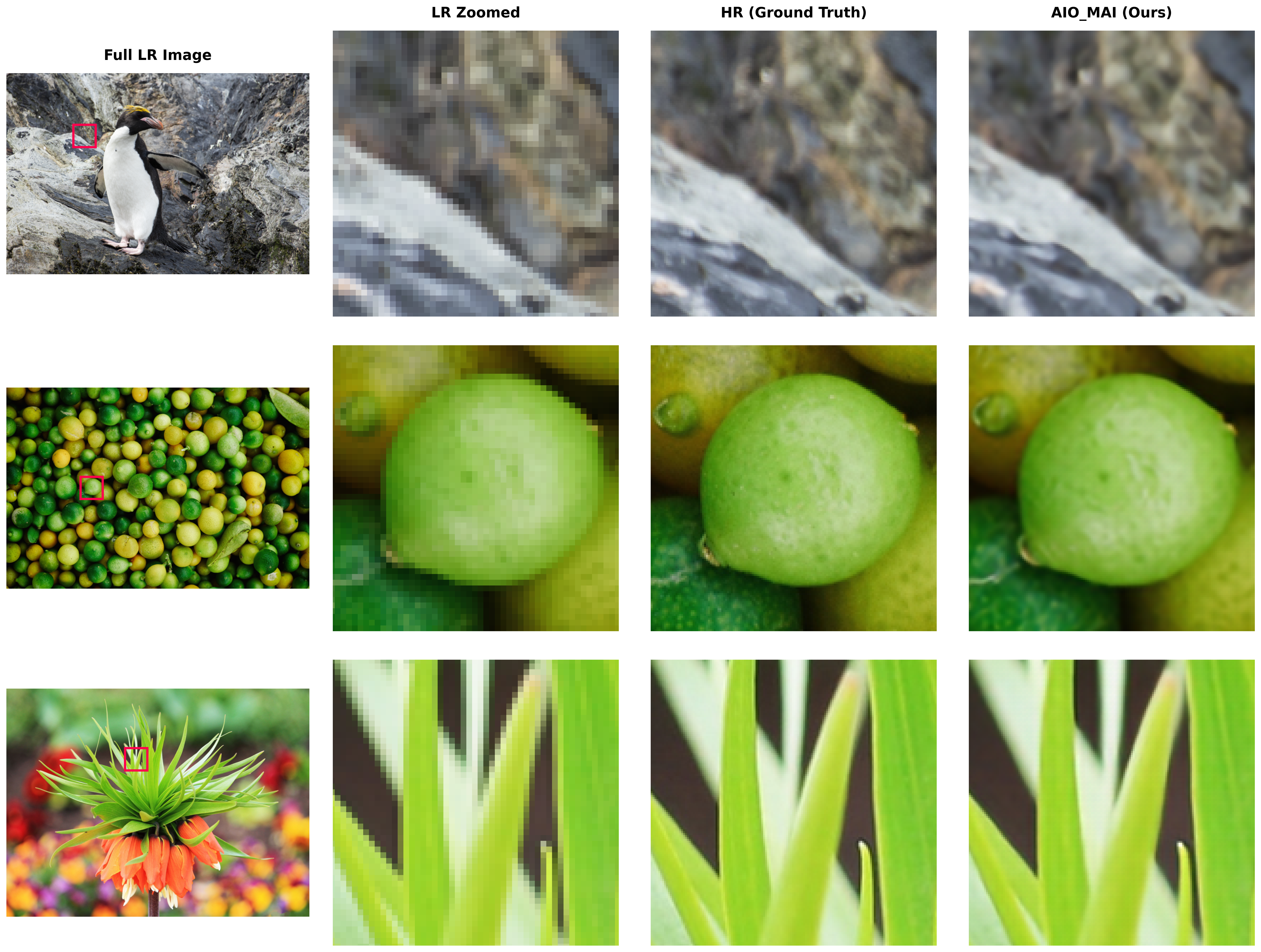}
\caption{Qualitative comparison of $\times3$ quantized image super-resolution on representative images from the DIV2K validation set. We compare our proposed method against the Bicubic baseline and the HR ground truth. Red boxes indicate the regions that are zoomed in for a detailed view. Despite strict INT8 quantization constraints, our AIO\_MAI model produces sharper edges and recovers finer textures, yielding results closer to the HR ground truth without obvious quantization artifacts.}
\label{fig:qualitative_comparison}
\end{figure*}

We provide a visual comparison against Bicubic interpolation and the High-Resolution (HR) ground truth in Fig.~\ref{fig:qualitative_comparison}. The proposed INT8 model reconstructs sharper edges, more regular structures, and richer high-frequency details than Bicubic, especially in regions with repeated patterns and thin boundaries, consistent with the quantitative gains.

This behavior is particularly important under challenge-style deployment constraints, where even a small accuracy drop after export can determine whether the final INT8 model remains above the required fidelity threshold. Taken together, these results show that deployment-oriented optimization should be treated as a first-class design objective rather than a post-hoc conversion step for lightweight super-resolution.

In the zoomed-in regions, Bicubic interpolation tends to produce blurry boundaries and oversmoothed textures, whereas our method restores clearer geometry and more faithful local details. The improvement is particularly visible in structured regions, where preserving regular patterns is important for perceptual quality, as well as in texture-rich areas that require better recovery of subtle high-frequency information. Compared with simple interpolation, the proposed method generates outputs that are visually closer to the HR reference.

Importantly, the final INT8 model remains visually stable after quantization, without obvious blocking artifacts or irregular edge distortions. This suggests that the proposed pipeline preserves visual fidelity effectively even under strict low-bit constraints. Overall, the qualitative results support the quantitative comparisons and show that the deployment-oriented design improves efficiency without sacrificing reconstruction quality.

\FloatBarrier

\subsection{Ablation Studies}

To evaluate the impact of our design choices, we conduct ablations on both the backbone structure and the quantization-oriented optimization strategy. These experiments help identify not only the strongest configuration, but also the components that are most important for stable low-bit $\times 3$ SR.

Table~\ref{tab:blocks} compares a RepVGG-style re-parameterized convolution block (RepConv)~\cite{ding2021repvgg}, our re-parameterizable depthwise block (RepDW), and a MobileOne-style block~\cite{vasu2023mobileone}. We include RepDW mainly as a comparison variant to examine whether a more aggressively factorized depthwise--pointwise design is suitable for compact and quantized SR. Specifically, RepDW replaces stronger convolutional mixing with depthwise multi-branch processing followed by pointwise projection, allowing us to test whether this lighter formulation can preserve reconstruction fidelity after quantization. In practice, however, RepDW performs clearly worse than both RepConv and MobileOne in FP32 and INT8 settings, and also suffers the largest quantization drop. By contrast, MobileOne achieves the best reconstruction quality after quantization, reaching 30.0003 dB PSNR and 0.856 SSIM, while also showing the smallest degradation from FP32 to INT8. These results indicate that the MobileOne-style block provides the most effective balance between train-time optimization strength, deploy-time simplicity, and quantization robustness, making it the best choice for our deployment-oriented SR model.

\begin{table}[htbp]
\centering
\caption{Comparison of different building blocks, including the performance drop from FP32 to the dynamic fully-quantized INT8 TFLite artifact.}
\label{tab:blocks}
\renewcommand{\arraystretch}{0.96}
\resizebox{0.96\linewidth}{!}{%
\begin{tabular}{lcccccc}
\toprule
\textbf{Block} & \multicolumn{2}{c}{\textbf{FP32}} & \multicolumn{2}{c}{\textbf{INT8}} & \multicolumn{2}{c}{\textbf{Drop} $\downarrow$} \\
\cmidrule(lr){2-3} \cmidrule(lr){4-5} \cmidrule(lr){6-7}
~ & \textbf{PSNR} & \textbf{SSIM} & \textbf{PSNR} & \textbf{SSIM} & \textbf{$\Delta$PSNR} & \textbf{$\Delta$SSIM} \\
\midrule
RepConv & 30.0897 & 0.855 & 29.8492  & 0.847 & 0.2405 & 0.008 \\
RepDW   & 29.3583 & 0.838 & 28.7031  & 0.814 & 0.6552 & 0.024 \\
\textbf{MobileOne} & \textbf{30.1350} & \textbf{0.859} & \textbf{30.0003} & \textbf{0.856} & \textbf{0.1347} & \textbf{0.003} \\
\bottomrule
\end{tabular}%
}
\renewcommand{\arraystretch}{1.0}
\end{table}

We further analyze the effect of the teacher network during Stage~3 QAT. As shown in Table~\ref{tab:ablation}, adding it improves convergence and raises INT8 PSNR from 29.91 dB to above 30.00 dB, highlighting the sensitivity of low-bit SR to quantization-aware stabilization. Although the numerical gain is modest, it is meaningful under a strict challenge threshold. This result indicates that stronger supervision remains useful even in the final quantized regime. This also suggests that teacher guidance helps reduce the optimization gap between floating-point training and deployed INT8 inference, leading to more reliable quantized reconstruction.

\begin{table}[htbp]
\centering
\caption{Ablation study on Stage 3 quantization-oriented optimization. Teacher-guided supervision improves quantized reconstruction, and its combination with the proposed QAT setting provides the most stable and effective INT8 performance.}
\label{tab:ablation}
\renewcommand{\arraystretch}{1.05}
\resizebox{1.08\linewidth}{!}{%
\begin{tabular}{lcccc}
\toprule
\textbf{Configuration} & \textbf{Teacher} & \textbf{Precision} & \textbf{PSNR (dB)} & \textbf{SSIM} \\
\midrule
Stage 3 (Direct QAT) & $\times$ & INT8 & 29.9114 & 0.853 \\
Stage 3 (Ours Full) & $\checkmark$ & INT8 & \textbf{30.0003} & \textbf{0.856} \\
\bottomrule
\end{tabular}%
}
\renewcommand{\arraystretch}{1.0}
\end{table}

\FloatBarrier
\section{Conclusion}
In this work, we presented a compact and deployment-oriented framework for ×3 single-image super-resolution under INT8 quantization constraints. Our method combines an extract–refine–upsample architecture with a MobileOne-style re-parameterizable backbone, PixelShuffle reconstruction, and a three-stage training pipeline that progressively aligns reconstruction fidelity with low-bit deployment. By combining Charbonnier and DCT losses, confidence-weighted distillation from a Mamba-based teacher, and QAT on the fused deploy graph, our method balances reconstruction quality, compactness, and quantization robustness. Experimental results show that the method remains competitive under mobile deployment constraints while preserving structural details and maintaining stable INT8 performance. These findings suggest that practical low-bit super-resolution benefits from jointly optimizing architecture, supervision, quantization stability, and deployment consistency. The results also show that a carefully designed lightweight model can remain effective after quantization without relying on large capacity or overly complex architectures. In future work, we plan to improve texture recovery and runtime efficiency through stronger lightweight backbones, better quantization strategies, and more hardware-aware optimization. Since the present evaluation is primarily centered on the official MAI 2026 target platform, broader validation on other mobile accelerators, such as MediaTek NPUs or Apple Neural Engine, remains an important direction for future work.

{
    \small
    \bibliographystyle{ieeenat_fullname}
    \bibliography{main}
}


\end{document}